\documentclass[sigconf]{acmart}
\AtBeginDocument{%
  \providecommand\BibTeX{{%
    \normalfont B\kern-0.5em{\scshape i\kern-0.25em b}\kern-0.8em\TeX}}}

\setcopyright{iw3c2w3g}
\copyrightyear{2021}
\acmYear{2021}

\acmConference[WWW '21]{Proceedings of the Web Conference 2021}{April 19--23, 2021}{Ljubljana, Slovenia}
\acmBooktitle{Proceedings of the Web Conference 2021 (WWW '21), April 19--23, 2021, Ljubljana, Slovenia}
\acmPrice{}
\acmDOI{10.1145/3442381.3449925}
\acmISBN{978-1-4503-8312-7/21/04}

\usepackage{graphicx}
\usepackage{subfigure}
\usepackage{amsmath}
\usepackage{amsfonts}
\usepackage{multirow}
\usepackage{comment}
\usepackage{hyperref}

\newcommand{\Modelshortsp}{KE-GCN }
\newcommand{\Modelshort}{KE-GCN}

\newtheorem{prop}{Proposition}

\begin{document}

\title{Knowledge Embedding Based Graph Convolutional Network}

\author{Donghan Yu}
\affiliation{%
  \institution{Carnegie Mellon University}
}
\email{dyu2@cs.cmu.edu}

\author{Yiming Yang}
\affiliation{%
  \institution{Carnegie Mellon University}
}
\email{yiming@cs.cmu.edu}

\author{Ruohong Zhang}
\affiliation{%
  \institution{Carnegie Mellon University}
}
\email{ruohongz@cs.cmu.edu}

\author{Yuexin Wu}
\affiliation{%
  \institution{Google}
}
\email{crickwu@google.com}

\renewcommand{\shortauthors}{Yu et al.}

\begin{abstract}
Recently, a considerable literature has grown up around the theme of Graph Convolutional Network (GCN).  How to effectively leverage the rich structural information in complex graphs, such as knowledge graphs with heterogeneous types of entities and relations, is a primary open challenge in the field. Most GCN methods are either restricted to graphs with a homogeneous type of edges (e.g., citation links only), or focusing on representation learning for nodes only instead of jointly propagating and updating the embeddings of both nodes and edges for target-driven objectives. This paper addresses these limitations by proposing a novel framework, namely the Knowledge Embedding based Graph Convolutional Network (KE-GCN), which combines the power of GCNs in graph-based belief propagation and the strengths of advanced knowledge embedding (a.k.a. knowledge graph embedding) methods, and goes beyond.  Our theoretical analysis shows that KE-GCN offers an elegant unification of several well-known GCN methods as specific cases, with a new perspective of graph convolution.
Experimental results on benchmark datasets show the advantageous performance of KE-GCN over strong baseline methods in the tasks of knowledge graph alignment and entity classification\footnote{Our code is publicly available in https://github.com/PlusRoss/KE-GCN}. 
\end{abstract}




\maketitle

\section{Introduction}
\label{intro}

Graph Convolution Networks (GCNs) have received increasing
attention in recent machine learning research due as powerful methods for graph-based node feature induction and belief propagation, 
and been successfully applied to many real-world problems,
including natural language processing~\cite{marcheggiani2017encoding}, computer vision~\cite{wang2018dynamic}, recommender
systems~\cite{ying2018graph}, epidemiological
forecasting~\cite{wu2018deep}, and more. 
Existing GCNs share the same core idea, i.e., using a graph to identify the neighborhood of each node, and to learn the embedding (vector representation) of that node via recursive aggregation of the neighborhood embeddings. In other words, graph convolution plays the central role in smoothing the learned representations (latent vectors) of nodes based on belief propagation over the entire graph. However, early work in GCNs also have a limitation in common, i.e., graph convolution is only used to learn node embedding conditioned on fixed edges~\cite{kipf2016semi}, instead of jointly learning the optimal embeddings for both nodes and edges. Later efforts move on the direction of learning of wights of edges in the input graph in parallel with the learning of node embedding~\cite{velivckovic2017graph,yu2019graph}, which is more powerful than early GCNs and more capable in model adaptation for downstream tasks. However, 
those GCNs have a constraint in common, that is, the edges in the input graph must be of homogeneous kind, such as the links in a citation graph or the elements in a co-occurrence count matrix. This constraint (or assumption) significantly limit the applicability of GCNs to a broad range of real-world applications where the capability to model \textit{heterogeneous} relations (edges) is crucial for the true utility of graph-based embedding and inference. 
Given that heterogeneous types of relations carry rich semantic information in KBs,  missing the ability of edge embedding fundamentally limits the expressiveness and prediction power of most GCNs. 

As an indirectly related area, methods for Knowledge Graph (KG) completion (a.k.a. knowledge graph embedding) have been intensively studied in recent years. Various algorithms have been developed for joint optimization of the embeddings of both entities and relations, with respect to the task of predicting unknown entity-relation-entity triplets based on observed triplets. Representative approaches include TransE~\cite{bordes2013translating}, DistMult~\cite{yang2014embedding}, ComplEx~\cite{trouillon2016complex}, RotatE~\cite{sun2019rotate}, QuatE~\cite{zhang2019quaternion} and more. The major difference of KG completion methods in contrast to GCNs is that the former do not explicitly leverage the belief propagation power of graph convolution in the representation learning process; instead, entity-relation-entity triplets are treated independently in their objective functions.  In other words, those methods lack the ability of using the graph structures to enforce the local/global smoothness in the embedding spaces for entities and relations.  

How to jointly leverage the strengths of both GCN models and KG completion methods for task-oriented representation learning of both entities and relations is an open challenge for research, which has not been studied in sufficient depth and is the focus of this paper. Representative works in this direction, or the only methods of this kind so far to our knowledge, are VR-GCN~\cite{ye2019vectorized}, TransGCN~\cite{cai2019transgcn}, and CompGCN~\cite{vashishth2019composition}. They use a graph neural network to jointly learn multi-layer latent representations (embeddings) for both entities and relations.  Specifically, 
the entity embedding process recursively aggregates both the neighborhood entity representations and relational representations, which makes sense; However, 
the \textit{relation} embedding part of the learning process leaves the \textit{entity} representations out of the picture,  
which is arguably sub-optimal and a fundamental limitation of those models. 

\begin{figure}[!tp]
    \centering
    \includegraphics[width=8cm]{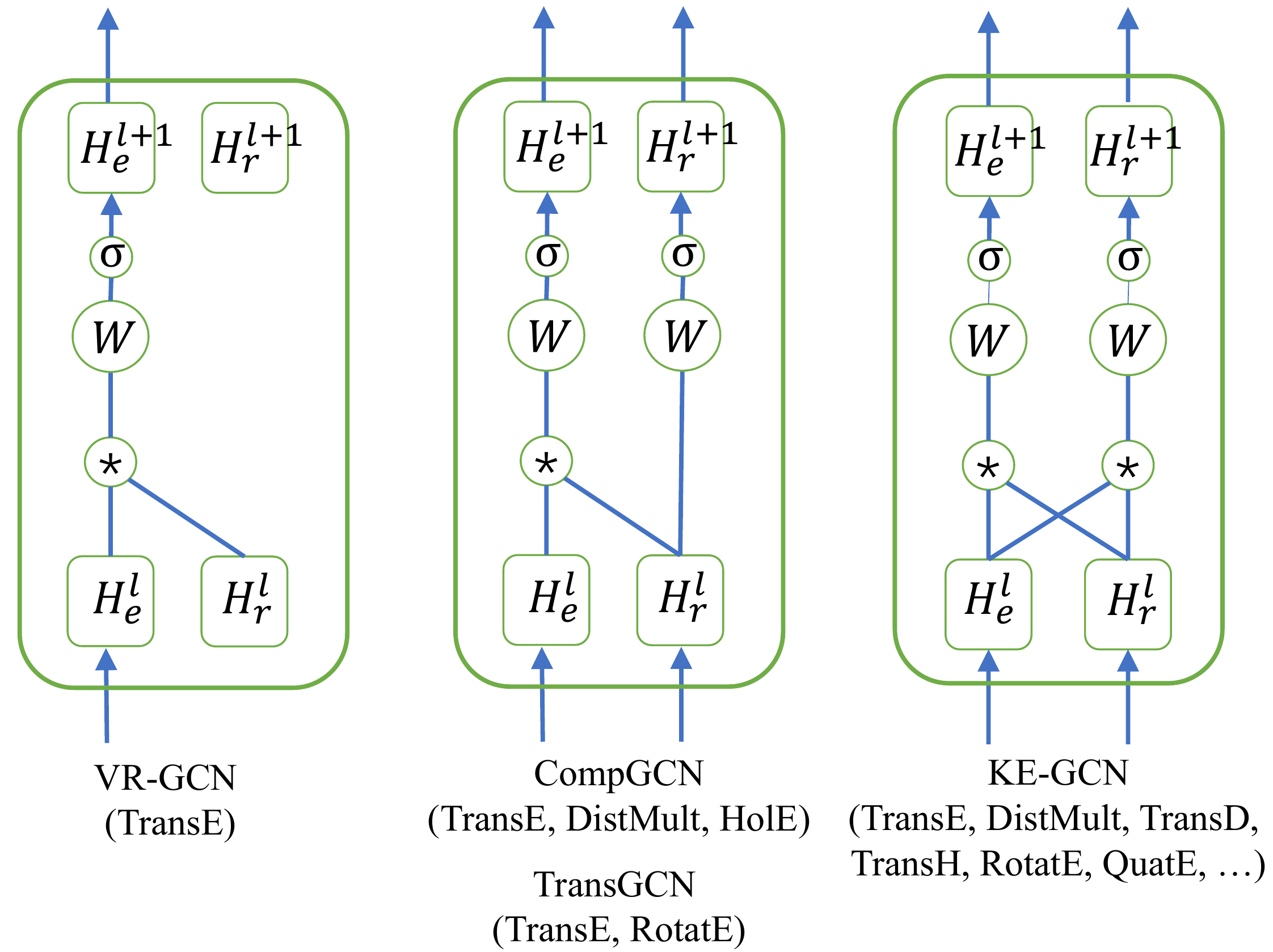}
    \caption{A simple realization of \Modelshortsp compared to previous works VR-GCN, TransGCN, and CompGCN. $H_e^l$ and $H_r^l$ means the entity (node) embedding and relation (edge) embedding at layer $l$ respectively. $\star$ denotes the graph convolution operation which aggregates the neighbour information. $W$ is the model parameter of linear transformation and $\sigma$ is the activation function. The names in the 
   brackets below correspond to the incorporated KG completion models.}
    \label{fig:model1}
\end{figure}

To address the aforementioned open challenge and the limitation of existing approaches, we propose a novel framework, namely \Modelshortsp (Knowledge Embedding based Graph Convolution Network).  It provides a theoretically sound generalization of existing GCN models, to allow the incorporation of various knowledge embedding methods for task-oriented embeddings of both entities and relations via graph convolution operations. Especially, in order to capture the rich semantics of heterogeneous relations in knowledge graphs, both entity embeddings and relation embeddings in our model are used to enforce optimization of each other in a recursive aggregation process. \autoref{fig:model1} illustrates the main differences between 
previous works and our model. A more detailed comparison will be provided in Section \ref{uni}. 
The contributions of our work can be summarized as follows:
\begin{itemize}
    \item We propose a novel framework \Modelshortsp which updates both entity and relation embeddings by graph convolution operation leveraging various knowledge embedding techniques.
    \item \Modelshortsp originates from a new intuition of GCN, and provides a unified view of several representative methods
    as its special and restricted cases. 
    \item Experiment results on benchmark datasets for knowledge graph alignment and entity classification tasks show that \Modelshortsp consistently and significantly outperforms other representative baseline methods.
\end{itemize}

We organize the rest of this paper as follows: Section \ref{rel} introduces related background on graph convolutional networks and knowledge graph embedding methods. Section \ref{model} describes our proposed framework and provides motivation of our method. Section \ref{uni} shows how our framework subsumes several representative approaches in a principled way. Section \ref{exp} reports our experimental results, followed by the conclusion in Section \ref{con}.

\section{Related Background}
\label{rel}

\subsection{Graph Convolutional Network}
Graph Convolutional Networks (GCNs) derive its idea from traditional Convolutional Neural Networks (CNNs) by extending convolutional operations onto non-Euclidean data structures. Early trials focus on spectral transformation over adjacency matrices~\cite{bruna2013spectral,defferrard2016convolutional} which requires tremendous computational cost of eigen decomposition. Recent work \cite{kipf2016semi} saves the computation of decomposition by making first order Chebyshev polynomial approximation and popularizes its use while keeping the strong performance over graph-structured data. This change formulates a uniform Message Passing framework for most followup works.

Thereafter, while many extension works tries to improve the effectiveness by reweighting edge weights \cite{velivckovic2017graph} or making residual links \cite{li2019deepgcns}, recent works try to leverage the power of GCNs on knowledge graph by allowing multi-relation edge types to interact in the overall framework \cite{vashishth2019composition,schlichtkrull2018modeling,shang2019end,HGNN,HGAT,KBAT,zhang2020relational}. These methods define new message passing routines by updating node representation using relation-specific transformation or interaction with relation embeddings. For example, KBAT \cite{KBAT} extends graph attention mechanisms to capture both entity and relation features in a multi-hop neighborhood of a given entity. RGHAT \cite{zhang2020relational} is equipped with relation-level attention and entity-level attention to calculate the weights for neighboring relations and entities respectively. CompGCN \cite{vashishth2019composition} leverages the entity-relation composition operations from KE methods like TransE \cite{bordes2013translating} to update entity embeddings. However, none of them have a symmetrical updating rule for their relation embedding, thus limiting the representation power of relation embeddings. Our method differs in that both entity embeddings and relation embeddings are utilized to update each other, where relation embeddings are also enhanced by neighbour context.

\subsection{Knowledge Embedding}
Traditional Knowledge Graph (KG) tasks focus on making link or entity predictions. These tasks mostly reduce to modeling (head entity, relation, tail entity) triplets. Many solutions~\cite{bordes2013translating,yang2014embedding,nickel2016holographic,wang2017knowledge} rely on learning embeddings for each part. Specifically, a scoring function $f$ is defined to measure the plausibility of triplets given the embeddings and help update the representation on the training data composed of positive and sampled negative triplets. By using different types of scoring functions, knowledge embedding methods can reflect different designing criteria, which include translation relation \cite{bordes2013translating,wang2014knowledge}, inner product \cite{yang2014embedding}, rotational relation \cite{sun2019rotate,zhang2019quaternion} and many others \cite{socher2013reasoning,dong2014knowledge}. For example, TransE \cite{bordes2013translating} assumes that given a triplet, the element-wise addition of head entity and relation embeddings should be close to the tail entity embedding. RotatE~\cite{sun2019rotate} is based on the assumption that relation serves as a rotation operation in the complex plane, and it claims to be the first model which can model and infer the following relational patterns: symmetry/anti-symmetry, inversion, and composition.

In our work, we borrow the idea of scoring function by adaptively allowing similar embedding learning design in the new GCN framework. We evaluate the performance of different scoring functions proposed in TransE~\cite{bordes2013translating}, DistMult~\cite{yang2014embedding},
TransH~\cite{wang2014knowledge},
TransD~\cite{ji2015knowledge},
RotatE~\cite{sun2019rotate}, and QuatE~\cite{zhang2019quaternion} through our framework on tasks including knowledge graph alignment and entity classification, which proves the effectiveness of our model design and such incorporation. A more detailed introduction about these methods will be provided in Section \ref{basicset}.

\section{Proposed Method}
\label{model}

\subsection{Reformulation of Vanilla GCN}

In vanilla GCN~\cite{kipf2016semi}, 
the multi-layer node embedding is updated as follows (we omit the normalization coefficient part for brevity):
\begin{align}
\mathbf{m}_v^{l+1} & = \sum_{u \in \mathcal{N}(v)} \mathbf{h}_u^l
\label{eq:gcn1}
\\
\mathbf{h}_v^{l+1}&= \sigma( W^{l} ( \mathbf{m}_v^{l+1} + \mathbf{h}_v^{l}))
\label{eq:gcn2}
\end{align}
where we denote by $\mathbf{h}_v^l$ the embedding of node $v$ at layer $l$,
by $\mathcal{N}(v)$ the set of immediate neighbours of node $v$,
by $\mathbf{m}_v^{l+1}$ the aggregated representation of those neighbors,
by $\sigma(\cdot)$ an activation function (e,g., element-wise sigmoid or ReLU), and by $W^l$ the matrix of model parameters to be learned by the GCN.
We can reformulate the vanilla GCN by introducing a scoring function $f$ that measures the plausibility of each edge. Edges observed in the graph tend to have higher scores than those that have not been observed. Specifically, for edge $(u,v)$, if we define $f$ as the inner product of the embeddings of the two connected nodes as  $f(\mathbf{h}_u,\mathbf{h}_v) = \mathbf{h}_u^T \mathbf{h}_v$,
then \autoref{eq:gcn1} have the equivalent forms of:
\begin{align}
\mathbf{m}_v^{l+1} = \sum_{u \in \mathcal{N}(v)}  \frac{\partial f(\mathbf{h}_u^l,\mathbf{h}_v^l)}{\partial \mathbf{h}_v^l}  =  \frac{\partial (\sum_{u \in \mathcal{N}(v)}  f(\mathbf{h}_u^l,\mathbf{h}_v^l))}{\partial \mathbf{h}_v^l}
\label{eq:gcn4}
\end{align}
It follows that $\mathbf{h}_v^{l} + \mathbf{m}_v^{l+1}$ can be regarded as one step gradient ascent to maximize the sum of scoring function $\sum_{u \in \mathcal{N}(v)}  f(\mathbf{h}_u^l,\mathbf{h}_v^l)$, with learning rate of $1$. Furthermore, \autoref{eq:gcn2} can be viewed as a generalized projection onto the embedding space for downstream tasks. 

The above reformulation provides an explicit view about \textit{what} the vanilla GCN is optimizing, instead of \textit{how} the updates are executed procedurally.  More importantly, it sheds light on how to design a more powerful framework to enable more generalized multi-relational graph convolution over knowledge graphs, which we introduce in the next section.

\subsection{The New Framework}

Since the relations in knowledge graphs are of heterogeneous types, we need to define the scoring function $f$ to measure the plausibility of entity-relation-entity triplets, instead of entity-entity pairs in vanilla GCN. Samely, triplets observed in the knowlede graph tend to have higher scores than those that have not been observed. For each triplet $(u,r,v)$, where $u,r,v$ denote head entity, relation, and tail entity respectively, the scoring value is calculated by $f(\mathbf{h}_u,\mathbf{h}_r,\mathbf{h}_v) \rightarrow \mathbb{R}$ using their embedding vectors. Note that most knowledge embedding techniques can be used to define $f$.

Analogous to \autoref{eq:gcn4}, if we denote $\mathbf{h}_v^l$ the embedding of entity $v$ at layer $l$, the entity updating rules are:
\begin{align}
\mathbf{m}_v^{l+1} & = \sum_{(u,r) \in \mathcal{N}_{\text{in}}(v)} W_{r}^{l} \frac{\partial f_{\text{in}}(\mathbf{h}_u^l, \mathbf{h}_r^l, \mathbf{h}_v^l)}{\partial \mathbf{h}_v^{l}}  \\
& + \sum_{(u,r) \in \mathcal{N}_{\text{out}}(v)} W_{r}^{l} \frac{\partial f_{\text{out}}(\mathbf{h}_v^l, \mathbf{h}_r^l, \mathbf{h}_u^l)}{\partial \mathbf{h}_v^{l}}
\label{model:e1}\\
\mathbf{h}_v^{l+1}&= \sigma_{\text{ent}}( \mathbf{m}_v^{l+1} + W_0^{l} \mathbf{h}_v^{l}) 
\label{model:e2}
\end{align}
where $\mathcal{N}_{\text{in}}(v) = \{(u,r) \mid u \stackrel{r}{\rightarrow} v\}$ is the set of immediate entity-relation neighbours of entity $v$ with an in-link while $\mathcal{N}_{\text{out}}(v) = \{(u,r) \mid u \stackrel{r}{\leftarrow} v\}$ is the set of immediate neighbours with an out-link from $v$. $\mathbf{h}_r^l$ means the embedding of relation $r$ at layer $l$. Notice that the linear transformation matrix $W_r^l$ is relation-specific, and the scoring functions $f_{\text{in}}$ are for the in-link neighbours and $f_{\text{out}}$ are for the out-link neighbours, respectively. $\sigma_{\text{ent}}(\cdot)$ denotes the activation function for entity update.

\begin{figure}[!tp]
    \centering
    \includegraphics[width=8.5cm]{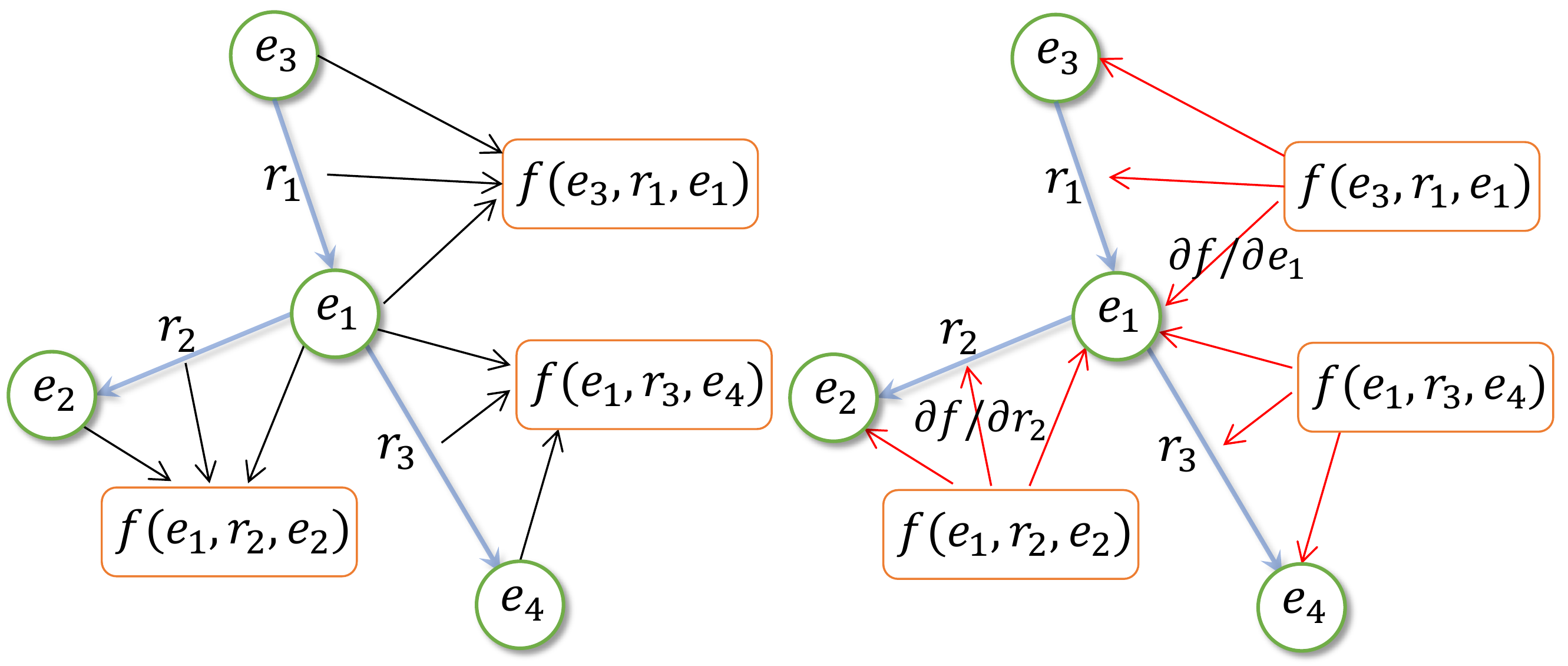}
    \caption{A demonstration for the message passing process of \Modelshort, where $e_1$, $e_2$, $e_3$, $e_4$ are entities and $r_1$, $r_2$, $r_3$ are relations in a knowledge graph. In one layer of our GCN model, it first calculate the plausibility score of each triplet by certain scoring function $f$, as shown in the left part, then pass the partial gradient (e.g. $\partial f / \partial e_1$) back to the corresponding entities and relations (e.g. $e_1$) to update their embeddings, as shown in the right part.}
    \label{fig:model2}
\end{figure}

The relation updating rules can be defined in a similar manner:
\begin{align}
\mathbf{m}_r^{l+1} & = \sum_{(u,v) \in \mathcal{N}(r)} \frac{\partial f_r(\mathbf{h}_u^l, \mathbf{h}_r^l, \mathbf{h}_v^l)}{\partial \mathbf{h}_r^{l}}
\label{model:r1}\\
\mathbf{h}_r^{l+1}&= \sigma_{\text{rel}}( W_{\text{rel}}^{l} ( \mathbf{m}_r^{l+1} +  \mathbf{h}_r^{l}) )
\label{model:r2}
\end{align}
where $\mathcal{N}(r) = \{ (u,v)\mid u \stackrel{r}{\rightarrow} v\}$ means the set of immediate entity neighbours of relation $r$, where the left side of tuple is head entity and the right side is tail entity. $\sigma_{\text{rel}}(\cdot)$ denotes the activation function for relation update. Figure \ref{fig:model2} provides a demonstration of our model. Note that our framework can subsume other representative methods, which will be introduced in the next section. 

Moreover, the derivative $\partial f/\partial \mathbf{h}$ in \autoref{model:e1} and \autoref{model:r1} can be calculated by \textbf{\textit{Auto Differentiation} (AD)} package of many existing libraries including Pytorch~\cite{paszke2017automatic} and Tensorflow~\cite{abadi2016tensorflow}, which makes our model easy to implement. Notice that this AD happens in the forward propagation process, which is different from the back-propagation during the training process. 

To apply our model on the downstream tasks, denoting the number of layers as $L$, we use the output entity embedding $\{ \mathbf{h}_v^{(L)}\}$ and relation embedding $\{ \mathbf{h}_r^{(L)} \}$ of the final layer to construct loss functions. For example, in entity classification task, we use cross-entropy loss based on entity labels; in knowledge graph alignment, we use the distance between the embedding vectors of two entities from different knowledge graphs for loss function. For details of the loss functions, please refer to Section \ref{sec:loss1} and \ref{sec:loss2}. The training manners are end-to-end in both tasks. In our experiments for these tasks, we set $f_{\text{in}} = f_{\text{out}} = f_r$ for simplicity and $W_{r}^{l} = W^{l}$ to prevent over-parameterization. Moreover, to avoid numerical instabilities
and exploding/vanishing gradients when used in a deep neural network model, \Modelshortsp applies the following normalization: we replace $\mathbf{m}_v^{l+1}$ in \autoref{model:e2} with $\alpha\mathbf{m}_v^{l+1}/(|\mathcal{N}_{\text{in}}(v)| + |\mathcal{N}_{\text{out}}(v)|)$ and $\mathbf{m}_r^{l+1}$ in \autoref{model:r2} with $\alpha\mathbf{m}_v^{l+1}/|\mathcal{N}(r)|$, where $|\cdot|$ means the cardinality of a set and $\alpha$ is a hyperparameter which controls the weight of aggregated neighbour information.

\section{Unified View of Representative Methods}
\label{uni}

In the following we provide a unified view of several representative GCN methods for knowledge graph modeling, by showing that they are restricted versions under our framework.

\subsection{CompGCN}
\label{compgcn}
CompGCN~\cite{vashishth2019composition} is most relevant to our method. In the $(l+1)$-th layer of CompGCN, the embeddings of each entity and relation are updated as: 
\begin{itemize}
    \item Entity update:
\begin{align}
\mathbf{m}_v^{l+1} & = \sum_{(u,r) \in \mathcal{N}_{\text{in}}(v)} W_{r}^{l} \phi_{\text{in}}\left(\mathbf{h}^{l}_{u}, \mathbf{h}^{l}_{r}\right) \notag \\ & + \sum_{(u,r) \in \mathcal{N}_{\text{out}}(v)} W_{r}^{l} \phi_{\text{out}}\left(\mathbf{h}^{l}_{u}, \mathbf{h}^{l}_{r}\right)\\
\mathbf{h}^{l+1}_{v}& =\sigma(\mathbf{m}_v^{l+1}+ W_0^l \mathbf{h}^{l}_{v} )
\end{align}
where $\phi_{\text{in}}, \phi_{\text{out}}: \mathbb{R}^{d_l} \times \mathbb{R}^{d_l} \rightarrow \mathbb{R}^{d_l}$ are the composition operators which can be element-wise subtraction, element-wise multiplication or circular-correlation~\cite{rabiner1975theory}. 
\item Relation update:
\begin{align}
\mathbf{h}_r^{l+1} & =W_{\text{rel}}^{l} \mathbf{h}_r^{l}
\end{align}
\end{itemize}

\begin{prop}
CompGCN can be fully recoverd by \Modelshortsp when 1) $f_{\text{in}}(\mathbf{h}^{l}_{u}, \mathbf{h}^{l}_{r},\mathbf{h}^{l}_{v}) = \phi_{\text{in}}\left(\mathbf{h}^{l}_{u}, \mathbf{h}^{l}_{r}\right)^T \mathbf{h}^{l}_{v}$, $f_{\text{out}}(\mathbf{h}^{l}_{v}, \mathbf{h}^{l}_{r},\mathbf{h}^{l}_{u})$ $=\phi_{\text{out}}\left(\mathbf{h}^{l}_{u}, \mathbf{h}^{l}_{r}\right)^T$ $\mathbf{h}^{l}_{v}$; and 2) $f_r = 0$; and 3) $\sigma_{\text{rel}}(\cdot)$ is the identity function.
\end{prop}

As shown above, in CompGCN, the relation embedding is only updated by linear transformation. While in \Modelshort, the relation representation update process aggregates neighbour entity representations, shown in Equation \ref{model:r1} and \ref{model:r2}, to capture the rich semantics of heterogeneous relations and learn better context-based relation embeddings. Additionally, our framework is more general since the scoring functions $f(\mathbf{h}_{u}, \mathbf{h}_{r},\mathbf{h}_{v})$ are not restricted to be $\phi\left(\mathbf{h}_{u}, \mathbf{h}_{r}\right)^T \mathbf{h}_{v}$, and other forms of knowledge graph embedding techniques such as TransH~\cite{wang2014knowledge}, TransD~\cite{ji2015knowledge}, MLP~\cite{dong2014knowledge}, and NTN~\cite{socher2013reasoning} can also be incorporated. Note that TransGCN \cite{cai2019transgcn} is very similar to CompGCN, and is also subsumed by our model.



\subsection{R-GCN}
R-GCN~\cite{schlichtkrull2018modeling} extends vanilla GCN with relation-specific linear transformations, without considering relation representations. The embedding update can be listed as follows:

\begin{itemize}
    \item Entity update:
\begin{align}
\mathbf{m}_v^{l+1} & = \sum_{(u,r) \in \mathcal{N}_{\text{in}}(v)} W_{r}^{l} \mathbf{h}^{l}_{u}  + \sum_{(u,r) \in \mathcal{N}_{\text{out}}(v)} W_{r}^{l}  \mathbf{h}^{l}_{u}\\
\mathbf{h}^{l+1}_{v}& =\sigma(\mathbf{m}_v^{l+1}+ W_0^l \mathbf{h}^{l}_{v} )
\end{align}
\item No relation update.
\end{itemize}

\begin{prop}
R-GCN can be fully recoverd by \Modelshortsp when 1) $f_{\text{in}}(\mathbf{h}^{l}_{u}, \mathbf{h}^{l}_{r},\mathbf{h}^{l}_{v}) =f_{\text{out}}(\mathbf{h}^{l}_{v}, \mathbf{h}^{l}_{r},\mathbf{h}^{l}_{u}) =(\mathbf{h}^{l}_{u})^T \mathbf{h}^{l}_{v}$; and 2) $\mathbf{h}_r^l = 0$ (no relation embedding).
\end{prop}

\subsection{W-GCN} 

W-GCN~\cite{shang2019end} treats the relation as learnable weights of edges, and applies vanilla GCN on the weighted simple graph. The update process can be written as:

\begin{itemize}
    \item Entity update:
\begin{align}
\mathbf{m}_v^{l+1} & = \sum_{(u,r) \in \mathcal{N}_{\text{in}}(v)} W^l (\alpha_{r}^{l} \mathbf{h}^{l}_{u})  + \sum_{(u,r) \in \mathcal{N}_{\text{out}}(v)} W^l (\alpha_{r}^{l}  \mathbf{h}^{l}_{u})\\
\mathbf{h}^{l+1}_{v}& =\sigma(\mathbf{m}_v^{l+1}+ W^l \mathbf{h}^{l}_{v} )
\end{align}
where $\alpha_{r}^{l} \in \mathbb{R}$ is a relation-specific learnable parameter.
\item No relation update.
\end{itemize}

\begin{prop}
W-GCN can be fully recoverd by \Modelshortsp when 1) $f_{\text{in}}(\mathbf{h}^{l}_{u}, \mathbf{h}^{l}_{r},\mathbf{h}^{l}_{v}) =f_{\text{out}}(\mathbf{h}^{l}_{v}, \mathbf{h}^{l}_{r},\mathbf{h}^{l}_{u}) =(\mathbf{h}^{l}_{u})^T \mathbf{h}^{l}_{v}$; and 2) $W_r^l = W^l \alpha_r^l$; and 3) $\mathbf{h}_r^l = 0$ (no relation embedding).
\end{prop}

\section{Experiments}
\label{exp}

\subsection{Basic Settings}
\label{basicset}

In this section, we conduct extensive experiments on two well-known tasks of knowledge graphs, \textit{graph alignment} and \textit{entity classification}, to demonstrate the effectiveness of our model. The computing infrastructure we use is NVIDIA RTX 2080Ti GPU in all the experiments. All the experiments are conducted by $5$ independent runs with different random seeds. In this paper, following previous works~\cite{schlichtkrull2018modeling,vashishth2019composition}, the input features of entities and relations are random vectors initialized by truncated normal distribution so that our model will rely solely on graph structure. We leave the combination of other features such as text description of entities and relations as future work. Our model is evaluated by combining the following representative knowledge graph embedding methods, with embedding of head entity, relation, and tail entity denoted as $\mathbf{h}_u$, $\mathbf{h}_r$, and $\mathbf{h}_v$ respectively. For each method, we show the corresponding scoring function in \Modelshortsp as follows:

\begin{itemize}
    \item TransE~\cite{bordes2013translating}: For $\mathbf{h}_u, \mathbf{h}_r, \mathbf{h}_v \in \mathbb{R}^{d}$,
    \begin{align}
    f(\mathbf{h}_u, \mathbf{h}_r, \mathbf{h}_v) = -\|\mathbf{h}_u+\mathbf{h}_r-\mathbf{h}_v\|_2^2.
    \end{align}
    
    \item DistMult~\cite{yang2014embedding}: For $\mathbf{h}_u, \mathbf{h}_r, \mathbf{h}_v \in \mathbb{R}^{d}$.
    \begin{align}
    f(\mathbf{h}_u, \mathbf{h}_r, \mathbf{h}_v) =\mathbf{h}_u^{T} \operatorname{diag}(\mathbf{h}_r) \mathbf{h}_v.
    \end{align}
    
    \item TransH~\cite{wang2014knowledge}: For $\mathbf{h}_u, \mathbf{h}_v \in \mathbb{R}^{d}, \mathbf{h}_r \in \mathbb{R}^{2d}$, and $\mathbf{h}_{r1}, \mathbf{h}_{r2} \in \mathbb{R}^{d}$,
    \begin{align}
    f(\mathbf{h}_u, \mathbf{h}_r, \mathbf{h}_v) & = -\|\mathbf{h}_u^{\prime}+\mathbf{h}_{r2}-\mathbf{h}_v^{\prime}\|_2^2, \\
    \mathbf{h}_u^{\prime} & = \mathbf{h}_u - \mathbf{h}_{r1}^T \mathbf{h}_{u} \mathbf{h}_{r1}, \\
    \mathbf{h}_v^{\prime} & = \mathbf{h}_v - \mathbf{h}_{r1}^T \mathbf{h}_{v} \mathbf{h}_{r1}, \\
    \mathbf{h}_r & = [\mathbf{h}_{r1} ; \mathbf{h}_{r2}],
    \end{align}
    where $[\cdot ; \cdot]$ means concatenation of two vectors.
    
    \item TransD~\cite{ji2015knowledge}: For $\mathbf{h}_u, \mathbf{h}_v, \mathbf{h}_r \in \mathbb{R}^{2d}$, and $\mathbf{h}_{u1}, \mathbf{h}_{u2}$, $\mathbf{h}_{v1}, \mathbf{h}_{v2}$, $\mathbf{h}_{r1}, \mathbf{h}_{r2} \in \mathbb{R}^{d}$,
    \begin{align}
        f(\mathbf{h}_u, \mathbf{h}_r, \mathbf{h}_v) & = -\|\mathbf{h}_u^{\prime}+\mathbf{h}_{r2}-\mathbf{h}_v^{\prime}\|_2^2 ,\\
        \mathbf{h}_u^{\prime} & = \mathbf{h}_{u1} + \mathbf{h}_{u2}^T \mathbf{h}_{u1} \mathbf{h}_{r1}, \\
        \mathbf{h}_v^{\prime} & = \mathbf{h}_{v1} - \mathbf{h}_{v2}^T \mathbf{h}_{v1} \mathbf{h}_{r1}, \\
        \mathbf{h}_u  = [\mathbf{h}_{u1} ; \mathbf{h}_{u2}], \mathbf{h}_v & = [\mathbf{h}_{v1} ; \mathbf{h}_{v2}], \mathbf{h}_r = [\mathbf{h}_{r1} ; \mathbf{h}_{r2}],
    \end{align}
    where $[\cdot ; \cdot]$ means concatenation of two vectors.
    
    \item RotatE~\cite{sun2019rotate}: For $\mathbf{h}_u, \mathbf{h}_r, \mathbf{h}_v \in \mathbb{C}^{d}$,
    \begin{align}
    f(\mathbf{h}_u, \mathbf{h}_r, \mathbf{h}_v) = -\|\mathbf{h}_u \circ \mathbf{h}_r-\mathbf{h}_v\|_2^2,
    \end{align}
    where $\circ$ denotes element-wise product and the modulus of any element in $\mathbf{h}_r$ is 1, i.e. $|\mathbf{h}_r[i]|=1 \ \forall i \in \{ 1,2,\cdots, d \}$. The norm of complex vector is defined as $\|\mathbf{v}\|_{p}=\sqrt[p]{\sum\left|\mathbf{v}_{i}\right|^{p}}$. 
    
    \item QuatE~\cite{zhang2019quaternion}: For $\mathbf{h}_u, \mathbf{h}_r, \mathbf{h}_v \in \mathbb{H}^{d}$,
    \begin{align}
    f(\mathbf{h}_u, \mathbf{h}_r, \mathbf{h}_v) = \mathbf{h}_u \otimes \mathbf{h}_r \cdot \mathbf{h}_v,
    \end{align}
    where $\otimes$ and $\cdot$ denote Hamilton Product and Inner Product in the hypercomplex space respectively.
\end{itemize}
Note that TransE and DistMult are widely used knowledge graph embedding techniques since they are simple and effective. RotatE and QuatE are recent proposed models and achieves state-of-the-art results in commonly used benchmark datasets for knowledge graph completion. TransH and TransD are the extensions of TransE, and can not be incorporated by previous works. We leave other embedding methods for future work.

\subsection{Knowledge Graph Alignment}

Knowledge graph alignment refer to the task which aims to find entities in two different knowledge graphs $KG_1$ and $KG_2$ that represent the same real-world entity. To apply our GCN framework on knowledge graph alignment, we follow the previous work~\cite{wang2018cross} which utilize two GCNs with shared parameters to model two knowledege graphs separately, then the output embeddings $\mathbf{h}$ of each final layer are used for entity alignment. To align from $KG_1$ to $KG_2$, for a specific entity/relation $u$ in $KG_1$, we compute the L1-distance between $u$ and each entity/relation $v$ in $KG_2$ using their embeddings $\mathbf{h}_{u}$ and $\mathbf{h}_{v}$, and returns a ranked list of entities/relations as candidate alignments based on the distances. The alignment can be also performed
from $KG_2$ to $KG_1$. In the experiments, we report the averaged results of both directions of KG alignment. 

\subsubsection{Loss Function}
\label{sec:loss1}
We denote the training entity alignment set as $S = \{ (u,v) \}$, where $u$ is the entity in $KG_1$ while entity $v$ belongs to $KG_2$ and they refer to the same real-world entity.
For the loss function, we follow previous work~\cite{wang2018cross} to use margin-based ranking loss:
\begin{align}
\mathcal{L}&= \sum_{(u, v) \in S} \sum_{(u^{\prime}, v^{\prime}) \in S_{(u, v)}^{\prime}} l(u,v,u^{\prime}, v^{\prime}) \\
l(u,v,& u^{\prime}, v^{\prime}) = [\|\mathbf{h}_{u} - \mathbf{h}_{v}\|_1+\gamma - \|\mathbf{h}_{u^{\prime}} - \mathbf{h}_{v^{\prime}}\|_1]_{+}
\end{align}
where $[x]_{+}=\max \{0, x\}$. $S_{(u, v)}^{\prime}$ denotes the set of negative entity alignments constructed by corrupting $(u,v)$, i.e. replacing $u$ or $v$ with a randomly chosen entity in $KG_1$ or $KG_2$. $\gamma$ means the margin hyper-parameter separating positive and negative entity alignments. Following \cite{wang2018cross}, we randomly choose 5 negtive alignments for each positive alignment and set $\gamma=3$.

\subsubsection{Dataset}

We use DBP15K~\cite{sun2017cross} which contains three datasets built from multi-lingual DBpedia, namely $\text{DBP}_{\text{ZH-EN}}$ (Chinese - English), $\text{DBP}_{\text{ZH-EN}}$ (Japanese - English) and $\text{DBP}_{\text{FR-EN}}$ (French - English) for knowledge graph alignment. A summary statistics of these datasets is provided in \autoref{data:ka}. Each dataset has $15,000$ reference entity alignment. Following previous work~\cite{wang2018cross,cao2019multi,sun2019knowledge}, we randomly split $30\%$ of them for training and the rest for testing. We further set aside $30\%$ of the training set as a validation set for hyperparameter tuning, and retrain the model on the whole training set to obtain test performance.

\begin{table}[!tp]
\caption{Dataset statistics of DBP15K for knowledge graph alignment task, including number of entities, relations, triplets of each knowledge graph.}
\centering
\begin{tabular}{lc|ccc}
\hline
\multicolumn{2}{c}{Datasets}                        & \#Entities & \#Relations & \#Triplets \\ \hline
\multicolumn{1}{c}{\multirow{2}{*}{$\text{DBP}_{\text{ZH-EN}}$}} & Chinese  & 66,469 & 2,830 & 153,929    \\
\multicolumn{1}{c}{}                     & English  & 98,125     & 2,317       & 237,674    \\ \hline
\multirow{2}{*}{$\text{DBP}_{\text{JA-EN}}$}                     & Japanese & 65,744     & 2,043       & 164,373    \\
 & English  & 95,680     & 2,096       & 233,319    \\ \hline
\multirow{2}{*}{$\text{DBP}_{\text{FR-EN}}$}                     & French   & 66,858     & 1,379       & 192,191    \\
& English  & 105,889    & 2,209       & 278,590    \\ \hline
\end{tabular}
\label{data:ka}
\end{table}

\subsubsection{Baselines}

To demonstrate the effectiveness of \Modelshort, we compare it with several representative multi-relational GCN baseline methods including R-GCN~\cite{schlichtkrull2018modeling}, W-GCN~\cite{shang2019end}, KBGAT~\cite{KBAT} and CompGCN \cite{vashishth2019composition}. We also include other baseline methods which are designed specifically for knowledge graph alignment task for a comprehensive comparison: MTransE~\cite{chen2017multilingual}, IPTransE~\cite{zhu2017iterative}, JAPE~\cite{sun2017cross}, AlignE~\cite{sun2018bootstrapping}, GCN-Align~\cite{wang2018cross}, MuGNN~\cite{cao2019multi}, and AliNet~\cite{sun2019knowledge}. Since our model solely relies on structure information, we do not compare with the alignment models which incorporate the surface information of entities into their representations~\cite{xu2019cross,wu2019relation}. Besides, some heterogeneous-graph-based models~\cite{HGAT,HGNN,GTN} which can not scale to knowledge graphs with hundreds or thousands of edge types are ruled out in comparison.

\begin{table*}[!tp]
\caption{Experiment results in knowledge graph entity alignment task on DBP15K datasets, where the average results over 5 different runs are reported. * indicate that results are directly taken from~\cite{sun2019knowledge}. The results of VR-GCN~\cite{ye2019vectorized} are directly taken from the original paper. CompGCN marked with $\dagger$ incorporates the composition operations in RotatE~\cite{sun2019rotate} and QuatE~\cite{zhang2019quaternion} while original CompGCN~\cite{vashishth2019composition} only contains subtraction, multiplication and circular-correlation operations.}
\centering
\setlength\tabcolsep{8pt} 
\begin{tabular}{lccccccccc}
\hline
\multirow{2}{*}{Models} & \multicolumn{3}{c}{$\text{DBP}_{\text{ZH-EN}}$} & \multicolumn{3}{c}{$\text{DBP}_{\text{JA-EN}}$} & \multicolumn{3}{c}{$\text{DBP}_{\text{FR-EN}}$} \\ \cmidrule(lr){2-4}\cmidrule(lr){5-7}\cmidrule(lr){8-10}
& MRR    & H@1    & H@10  & MRR    & H@1    & H@10  & MRR    & H@1    & H@10  \\ \hline
MTransE\textsuperscript{$*$}\cite{chen2017multilingual}                 & 0.364  & 30.8  & 61.4 & 0.349  & 27.9  & 57.5 & 0.335  & 24.4  & 55.6 \\
IPTransE\textsuperscript{$*$}\cite{zhu2017iterative}                & 0.516  & 40.6  & 73.5 & 0.474  & 36.7  & 69.3 & 0.451  & 33.3  & 68.5 \\
JAPE\textsuperscript{$*$}\cite{sun2017cross}                    & 0.490   & 41.2  & 74.5 & 0.476  & 36.3  & 68.5 & 0.430   & 32.4  & 66.7 \\
AlignE\textsuperscript{$*$}\cite{sun2018bootstrapping}                  & 0.581  & 47.2  & 79.2 & 0.563  & 44.8  & 78.9 & 0.599  & 48.1  & 82.4 \\
GCN-Align\textsuperscript{$*$}\cite{wang2018cross}               & 0.549  & 41.3  & 74.4 & 0.546  & 39.9  & 74.5 & 0.532  & 37.3  & 74.5 \\
MuGCN\textsuperscript{$*$}\cite{cao2019multi}                   & 0.611  & 49.4  & \textbf{84.4} & 0.621  & 50.1  & \textbf{85.7} & 0.621  & 49.5  & 87.0  \\
AliNet\textsuperscript{$*$}\cite{sun2019knowledge}                  & 0.628  & 53.9  & 82.6 & 0.645  & 54.9  & 83.1 & 0.657  & 55.2  & 85.2 \\ \hline
R-GCN\textsuperscript{$*$}\cite{schlichtkrull2018modeling}                   & 0.564  & 46.3  & 73.4 & 0.571  & 47.1  & 75.4 & 0.570   & 46.9  & 75.8 \\
W-GCN~\cite{shang2019end}                   &   0.553     &    43.6    &   73.8    & 0.554       &   41.2     &  74.7     & 0.541       &   39.8     &   74.4    \\
VR-GCN~\cite{ye2019vectorized}                   &   0.501 & 38.0 & 73.3 & 0.470 & 35.2 & 72.2 & 0.495 & 36.1 & 75.1   \\
KBGAT~\cite{KBAT}                   &  0.582 & 48.0 & 77.3 & 0.582 & 47.6 & 77.7 & 0.593 & 47.4 &  80.9  \\
CompGCN\cite{vashishth2019composition}                 &    0.605   &  49.4     &  81.2        &    0.614    &   50.4    &    82.2    & 0.625 & 50.5      &   85.0    \\ 
CompGCN\textsuperscript{$\dagger$}             &    0.628   &  52.8     &  81.1        &    0.629    &   52.8    &    81.5    &    0.641    &  52.6 &  85.4    \\ \hline
\Modelshort                 &  \textbf{0.664}      &  \textbf{56.2}   &  84.2     &    \textbf{0.670}    &   \textbf{57.0}     & 85.2       &     \textbf{0.683}  &   \textbf{57.2}      &   \textbf{88.5}    \\ \hline
\end{tabular}
\label{exp:ka}
\end{table*}

\subsubsection{Implementation Detail}
After performing grid search for hyperparameters, we set the learning rate to be 0.01, the hidden dimension to be 200, the number of layers to be 4, and $\alpha = 0.3$. The activation function is set as ReLU. We train our model in full-batch setting using Adam~\cite{kingma2014adam}. Following previous work~\cite{cao2019multi,sun2019knowledge}, we report Hits@1, Hits@10, Mean Reciprocal Rank (MRR) to evaluate the entity alignment performance.

\subsubsection{Results of Entity Alignment}

\autoref{exp:ka} shows the experiment results in DBP15K datasets comparing with all the baseline methods. Our model achieves the best or highly competitive results in all the three datasets, outperforming the baseline methods by a large margin. Specifically, our model outperforms the best baseline methods by 5.7\%, 6.5\%, and 6.6\% in MRR on $\text{DBP}_{\text{ZH-EN}}$, $\text{DBP}_{\text{JA-EN}}$, and $\text{DBP}_{\text{FR-EN}}$ respectively. Note that even equipped with the recent state-of-the-art knowledge graph embedding methods, i.e., RotatE~\cite{sun2019rotate} and QuatE~\cite{zhang2019quaternion}, CompGCN~\cite{vashishth2019composition} still obtains much lower performance than \Modelshort.
A more clear comparison is shown in \autoref{exp:ka3}, where we present the results on  $\text{DBP}_{\text{ZH-EN}}$ dataset of CompGCN and \Modelshortsp incorporated with TransE and QuatE. 
The results demonstrate the effectiveness of graph convolution updating for relation embeddings. 

We also report the performance of our proposed model with different knowledge graph embedding techniques including TransE, DistMult,
TransH,
TransD,
RotatE, and QuatE on $\text{DBP}_{\text{ZH-EN}}$, $\text{DBP}_{\text{JA-EN}}$, and $\text{DBP}_{\text{FR-EN}}$ datasets as shown in \autoref{exp:ka2}, \autoref{exp:ka22}, and \autoref{exp:ka23} respectively. Note that the choice of embedding techniques does have a large impact on the performance, and QuatE achieves the best results, which is reasonable since it also outperforms other methods in knowledge graph completion task, and satisfies the essential of relational representation learning
(i.e., modeling symmetry, anti-symmetry and inversion relation).

\begin{table}[!tp]
\caption{Direct comparison of CompGCN and \Modelshortsp incoporated with the same knowledge graph embedding techniques (TransE~\cite{bordes2013translating} and QuatE~\cite{zhang2019quaternion}) on $\text{DBP}_{\text{ZH-EN}}$.}
\label{exp:ka3}
\centering
\setlength{\tabcolsep}{1.2mm}{
\begin{tabular}{lccc}
\hline
Model             & MRR & H@1 & H@10 \\ \hline
CompGCN (TransE) & 0.605 $\pm$ 0.003 & 49.4 $\pm$ 0.4     & 81.2 $\pm$ 0.3     \\
\Modelshortsp (TransE)      & \textbf{0.648 $\pm$ 0.003}   & \textbf{54.3 $\pm$ 0.3}      &  \textbf{83.4 $\pm$ 0.3}      \\ \hline
CompGCN (QuatE) &  0.628 $\pm$ 0.003  &  52.8 $\pm$ 0.3      &  81.1 $\pm$ 0.4    \\
\Modelshortsp (QuatE)   &  \textbf{0.664 $\pm$ 0.004}   &  \textbf{56.2 $\pm$ 0.4}     & \textbf{84.2 $\pm$ 0.4}    \\ \hline
\end{tabular}
}
\end{table}

\begin{table}[!tp]
\caption{Knowledge graph entity alignment results over 5 different runs on $\text{DBP}_{\text{ZH-EN}}$ by incorporating different knowledge graph embedding methods into our model.}
\label{exp:ka2}
\centering
\begin{tabular}{lccc}
\hline
\Modelshortsp (X)             & MRR & H@1 & H@10 \\ \hline
X = TransE   & 0.648 $\pm$ 0.003   & 54.3 $\pm$ 0.3      &  83.4 $\pm$ 0.3    \\
X = TransH   &   0.650 $\pm$ 0.003      &  54.3 $\pm$ 0.4   &    \textbf{84.4 $\pm$ 0.3}  \\
X = DistMult &  0.621 $\pm$ 0.003       &  52.0 $\pm$ 0.4   & 80.3 $\pm$ 0.4     \\
X = TransD &   0.635 $\pm$ 0.003    &  53.1 $\pm$ 0.3  &     82.7 $\pm$ 0.4\\
X = RotatE   &  0.653 $\pm$ 0.004   & 54.9 $\pm$ 0.4      &  83.8 $\pm$ 0.4   \\
X = QuatE   &  \textbf{0.664 $\pm$ 0.004}   &  \textbf{56.2 $\pm$ 0.4}     & 84.2 $\pm$ 0.4  \\\hline  
\end{tabular}
\end{table}

\begin{table}[htp]
\caption{Knowledge graph entity alignment results over 5 different runs on $\text{DBP}_{\text{JA-EN}}$ dataset by incorporating different knowledge graph embedding methods into our model.}
\label{exp:ka22}
\centering
\begin{tabular}{lccc}
\hline
\Modelshortsp (X)             & MRR & H@1 & H@10 \\ \hline
X = TransE   & 0.652 $\pm$ 0.003   & 54.8 $\pm$ 0.3      &  84.3 $\pm$ 0.3    \\
X = TransH   &   0.654 $\pm$ 0.003      &  54.6 $\pm$ 0.4   &    \textbf{85.5 $\pm$ 0.3}  \\
X = DistMult &  0.622 $\pm$ 0.004       &  51.7 $\pm$ 0.5   & 81.2 $\pm$ 0.3     \\
X = TransD &   0.652 $\pm$ 0.001    &  54.5 $\pm$ 0.2  &     85.4 $\pm$ 0.2\\
X = RotatE   &  0.659 $\pm$ 0.002   & 55.7 $\pm$ 0.2      &  85.0 $\pm$ 0.3   \\
X = QuatE   &  \textbf{0.670 $\pm$ 0.001}   &  \textbf{57.0 $\pm$ 0.2}     & 85.2 $\pm$ 0.3  \\\hline  
\end{tabular}
\end{table}

\begin{table}[htp]
\caption{Knowledge graph entity alignment results over 5 different runs on $\text{DBP}_{\text{FR-EN}}$ dataset by incorporating different knowledge graph embedding methods into our model.}
\label{exp:ka23}
\centering
\begin{tabular}{lccc}
\hline
\Modelshortsp (X)             & MRR & H@1 & H@10 \\ \hline
X = TransE   & 0.669 $\pm$ 0.002   & 55.9 $\pm$ 0.2      &  87.5 $\pm$ 0.2    \\
X = TransH   &   0.673 $\pm$ 0.002      &  56.1 $\pm$ 0.2   &    87.7 $\pm$ 0.2  \\
X = DistMult &  0.640 $\pm$ 0.002       &  52.4 $\pm$ 0.2   & 84.7 $\pm$ 0.2     \\
X = TransD &   0.660 $\pm$ 0.002    &  54.2 $\pm$ 0.2  &     87.6 $\pm$ 0.1\\
X = RotatE   &  0.673 $\pm$ 0.002   & 56.0 $\pm$ 0.3      &  88.2 $\pm$ 0.2   \\
X = QuatE   &  \textbf{0.683 $\pm$ 0.002}   &  \textbf{57.2 $\pm$ 0.3}     & \textbf{88.5 $\pm$ 0.2}  \\\hline  
\end{tabular}
\end{table}

\subsubsection{Results of Relation Alignment}

We also test on relation alignment task to demonstrate the importance of our proposed relation embedding update process. Since the number of reference aligned relations is very small, we train the model on the entity alignment task mentioned above and directly use the trained relation embedding for relation alignment (as zero-shot evaluation). The results over 5 different runs are shown in \autoref{exp:ka4}, where we see that our model significantly outperforms CompGCN~\cite{vashishth2019composition} and VR-GCN~\cite{ye2019vectorized}. This demonstrate the importance of incorporating entity representation into the update of relation embeddings. We do not compare with R-GCN~\cite{schlichtkrull2018modeling} and W-GCN~\cite{shang2019end} since relation embeddings are not involved in their models.

\begin{table}[!tp]
\caption{Knowledge graph relation alignment results over 5 different runs on DBP15K datasets. All the models are incorporated with the same KG completion method TransE~\cite{bordes2013translating}.}
\label{exp:ka4}
\centering
\begin{tabular}{lccc}
\hline
Model             & $\text{DBP}_{\text{ZH-EN}}$ & $\text{DBP}_{\text{JA-EN}}$ & $\text{DBP}_{\text{FR-EN}}$ \\ \hline
VR-GCN  &  0.352  $\pm$ 0.006      & 0.335  $\pm$ 0.008 &  0.280  $\pm$ 0.017  \\
KBGAT  &  0.341  $\pm$ 0.010      & 0.330  $\pm$ 0.013 &  0.274  $\pm$ 0.022  \\
 CompGCN  &   0.366 $\pm$ 0.007      &  0.347  $\pm$ 0.009    & 0.284  $\pm$ 0.015   \\
\Modelshortsp &   \textbf{0.514  $\pm$ 0.006}      &  \textbf{0.466  $\pm$ 0.011}   & \textbf{0.412  $\pm$ 0.021}     \\\hline  
\end{tabular}
\end{table}

\subsection{Knowledge Graph Entity Classification}

Entity Classification is the task of predicting the labels of entities in a given knowledge graph. We follow previous work~\cite{schlichtkrull2018modeling,vashishth2019composition} to use the entity output of the last layer in \Modelshortsp for label classification.

\subsubsection{Loss Function}
\label{sec:loss2}
In this paper we conduct experiments on both multi-class classification and multi-label classification. Denoting $N$ as the number of entities, $C$ as the number of classes, and $\mathcal{Y}_{L}$ as the set of entity indices that have labels. For multi-class classification, we use the following loss:
\begin{align}
\mathcal{L}=-\sum_{u \in \mathcal{Y}_{L}} \sum_{c=1}^{C} Y_{uc} \ln \widehat{Y}_{uc}
\end{align}
where $Y_{uc} = 1$ means that the true class of entity $u$ is $c$, otherwise $Y_{uc} = 0$. $\widehat{Y} \in \mathbb{R}^{N \times C}$ is the output of GCN model, which comes after a row-wise softmax function. For multi-label classification, we use:
\begin{align}
\mathcal{L} =  -\sum_{u \in \mathcal{Y}_{L}} \sum_{c=1}^{C} \left[ Y_{uc} \ln \widehat{Y}_{uc} \right.  \left. + (1-Y_{uc})\ln (1-\widehat{Y}_{uc})\right]
\end{align}
$Y_{uc} = 1$ means that entity $u$ contains label $c$, otherwise $Y_{uc} = 0$. $\widehat{Y} \in \mathbb{R}^{N \times C}$ is the output of GCN model, which comes after an element-wise sigmoid function.

\subsubsection{Dataset}
We conduct experiments on the following datasets: AM~\cite{ristoski2016collection} which contains relationship between different artifacts in Amsterdam Museum, WN~\cite{bordes2013translating,tu2018structural} which consists of a collection of triplets (synset, relation, synset) extracted from WordNet 3.0~\cite{miller1995wordnet}, FB15K~\cite{bordes2013translating,xie2016representation} which is extracted from a typical large-scale knowledge graph Freebase~\cite{bollacker2008freebase}. The statistics of these datasets is shown in \autoref{data:kc}. Each entity in AM and WN datasets has at most one label while entities in FB15K can have multiple labels\footnote{Please refer to \cite{tu2018structural} and \cite{xie2016representation} for details of label collection in WN and FB15K datasets respectively.}. For AM dataset, we follow the train/test split convention~\cite{ristoski2016collection,schlichtkrull2018modeling}. As for WN and FB15K datasets, we randomly split the labeled entities into train/valid/test by the ratio of 10\%/10\%/80\%.

\begin{table}[!tp]
\caption{Number of entities, relations, edges and classes along with the number of labeled entities for each dataset in entity classification task. Labeled denotes the subset of entities that have labels and that are to be classified. Classes denotes the total number of categories of labels.}
\label{data:kc}
\centering
\begin{tabular}{lrrr}
\hline
Datasets    & AM        & WN & FB15K  \\\hline
\#Entities  & 1,666,764 & 40,551  & 14,904 \\
\#Relations & 133       & 18      & 1,341  \\
\#Triplets  & 5,988,321 & 145,966 & 579,654 \\
\#Labeled   & 1,000     & 31,943  & 13,445 \\
\#Classes   & 11        &   24      & 50 \\ \hline
\end{tabular}
\vspace{-3mm}
\end{table}


\subsubsection{Baselines}
We compare \Modelshortsp with vanilla GCN~\cite{kipf2016semi} and the relation-based GCN models including R-GCN~\cite{schlichtkrull2018modeling}, W-GCN~\cite{shang2019end}, KBGAT~\cite{KBAT}, and CompGCN~\cite{vashishth2019composition}. For vanilla GCN, we transform the multi-relational graphs to homogenous graphs, by setting the weight of edge between two entities as the number of their relations. 

\subsubsection{Implementation Detail}
After performing grid search for hyperparameters, we set learning rate to be 0.01, hidden dimension to be 32, number of layers to be 4, and $\alpha = 0.3$. The activation function is set as ReLU. We train our model in full-batch setting using Adam~\cite{kingma2014adam}. In AM and WN, Accuracy are reported to evaluate the entity classification performance. While in FB15K, we report Precision@1 (P@1), Precision@5 (P@5) and NDCG@5 (N@5). 

\subsubsection{Results}

The experiment results over 5 different runs on AM and WN datasets are shown in \autoref{exp:kc1}, where the average of classification accuracy is reported. The results on FB15K dataset under metrics of P@1, P@5, N@5 are shown in \autoref{exp:kc2}. From these results, \Modelshortsp outperforms all the baseline GCN methods, which demonstrate the effectiveness of our proposed model in entity classification task, including multi-class classification and multi-label classification. We report the best results of CompGCN~\cite{vashishth2019composition} and \Modelshortsp incorporated with different knowledge graph embedding methods, where combining TransE~\cite{bordes2013translating} achieves the highest performance. The results of KE-GCN incorporated with different KE methods are shown in Table \ref{exp:akc1} and Table \ref{exp:akc2}. Additionally, we see that vanilla GCN performs worse than any relational GCNs in all the datasets, which demonstrates that relation modelling is significant for entity classification.

\begin{table}[!tp]
\caption{The mean and standard deviation of classification accuracy over 5 different runs on AM and WN datasets for multi-class classification task. * indicates the results that are directed taken from~\cite{vashishth2019composition}.}
\centering
\begin{tabular}{lcc}
\hline
Models    & AM        & WN  \\\hline
GCN  & 86.2 $\pm$ 1.4 & 53.4 $\pm$ 0.2   \\
R-GCN &   89.3\textsuperscript{*}     & 55.1 $\pm$ 0.6       \\
W-GCN &  90.2 $\pm$ 0.9\textsuperscript{*}      &  54.2 $\pm$ 0.5 \\
KBGAT &  85.7 $\pm$ 1.7      &  53.7 $\pm$ 1.1 \\
CompGCN  & 90.6 $\pm$ 0.2\textsuperscript{*} & 55.9 $\pm$ 0.4 \\
\Modelshort   &   \textbf{91.2 $\pm$ 0.2}  & \textbf{57.8 $\pm$ 0.5}       \\ \hline
\end{tabular}
\label{exp:kc1}
\end{table}

\begin{table}[!tp]
\caption{The mean and standard deviation of Precision@1 (P@1), Precision@5 (P@5), NDCG@5 (N@5) over 5 different runs on FB15K dataset for multi-label classification task. }
\centering
\begin{tabular}{lccc}
\hline
Models    & P@1 & P@5 & N@5  \\\hline
GCN  & 86.1 $\pm$ 0.3 & 69.0 $\pm$ 0.3 & 82.7 $\pm$ 0.2  \\
R-GCN & 91.7 $\pm$ 0.6 & 73.0 $\pm$ 0.4 &   89.5 $\pm$ 0.6       \\
W-GCN & 91.2 $\pm$ 0.6 & 72.8 $\pm$ 0.3 &   88.6 $\pm$ 0.5   \\
KBGAT & 90.5 $\pm$ 0.7 & 72.4 $\pm$ 0.8 &   87.5 $\pm$ 0.8   \\
CompGCN  & 92.5 $\pm$ 0.1 & 74.0 $\pm$ 0.3 &   90.1 $\pm$ 0.2   \\
\Modelshort   & \textbf{94.3 $\pm$ 0.2} & \textbf{74.7 $\pm$ 0.2} &   \textbf{91.6 $\pm$ 0.2}     \\ \hline
\end{tabular}
\label{exp:kc2}
\end{table}

\begin{table}[htp]
\caption{Entity classification accuracy results over 5 different runs on AM and WN datasets by incorporating different knowledge graph embedding methods into our model.}
\centering
\begin{tabular}{lcc}
\hline
\Modelshortsp (X)       & AM        & WN  \\\hline
X = TransE   & \textbf{91.2 $\pm$ 0.2}   & \textbf{57.8 $\pm$ 0.5}         \\
X = TransH   &   90.5 $\pm$ 0.3      &  57.4 $\pm$ 0.3     \\
X = DistMult &  89.5 $\pm$ 0.4       &  56.4 $\pm$ 0.1        \\
X = TransD &   90.1 $\pm$ 0.2    &  57.1 $\pm$ 0.2  \\
X = RotatE   &  90.6 $\pm$ 0.4  & 56.6 $\pm$ 0.3         \\
X = QuatE   &  91.0 $\pm$ 0.4   &  56.9 $\pm$ 0.3      \\\hline  
\end{tabular}
\label{exp:akc1}
\end{table}

\begin{table}[htp]
\caption{Entity classification results over 5 different runs on FB15K dataset by incorporating different knowledge graph embedding methods into our model.}
\centering
\begin{tabular}{lccc}
\hline
\Modelshortsp (X)    & P@1 & P@5 & N@5  \\\hline
X = TransE   & 94.3 $\pm$ 0.2   & \textbf{74.7 $\pm$ 0.2}      &  \textbf{91.6 $\pm$ 0.2}    \\
X = TransH   &   94.3 $\pm$ 0.3      &  74.6 $\pm$ 0.2   &    \textbf{91.6 $\pm$ 0.2}  \\
X = DistMult &  94.2 $\pm$ 0.1       &  72.4 $\pm$ 0.4   & 89.9 $\pm$ 0.3     \\
X = TransD &   93.9 $\pm$ 0.4   &  74.5 $\pm$ 0.2  &     91.3 $\pm$ 0.2\\
X = RotatE   &  93.8 $\pm$ 0.4   & 73.7 $\pm$ 0.2      &  90.5 $\pm$ 0.2   \\
X = QuatE   &  \textbf{94.9 $\pm$ 0.4}   &  74.2 $\pm$ 0.3     & \textbf{91.6 $\pm$ 0.2}  \\\hline 
\end{tabular}
\label{exp:akc2}
\end{table}

\section{Conclusion}
\label{con}

This paper presents a novel framework which leverages various knowledge graph embedding methods into GCNs for multi-relational graph modelling, and more importantly, update both entity and relation representation using graph convolution operation. We show that our model originates from a new intuition behind graph convolution in the view of generalized projected gradient ascent, and subsumes other representative methods as its special and restricted cases. 
Experiments show that the proposed model obtains the state-of-the-art results in benchmark datasets of two well-known tasks: knowledge graph alignment and entity classification. In the future, we plan to apply our framework into a broader range of domains containing knowledge graphs, including Q\&A, recommender system, computer vision and time series analysis. It's also worth exploring to go beyond triplets and extend our framework for knowledge hypergraphs.


\begin{acks}
We thank the reviewers for their helpful comments. This work is supported in part by the National Science Foundation (NSF) under grant IIS-1546329, and by the United States Department of Energy via the Brookhaven National Laboratory under Contract No. 384608.
\end{acks}

\bibliographystyle{ACM-Reference-Format}
\bibliography{main}









\end{document}